\documentclass{article} 
\usepackage{iclr2022_conference,times}

\usepackage{amsmath,amsfonts,bm}

\def\eqref#1{equation~\ref{#1}}

\def\1{\bm{1}}

\DeclareMathAlphabet{\mathsfit}{\encodingdefault}{\sfdefault}{m}{sl}
\SetMathAlphabet{\mathsfit}{bold}{\encodingdefault}{\sfdefault}{bx}{n}

\usepackage[utf8]{inputenc} 
\usepackage[T1]{fontenc}    
\usepackage{hyperref}       
\usepackage{url}            
\usepackage{booktabs}       
\usepackage{amsfonts}       
\usepackage{nicefrac}       
\usepackage{microtype}      
\usepackage{xcolor}         

\usepackage{bm}
\usepackage{bbm}
\usepackage{mathtools}
\usepackage{wrapfig}
\usepackage{tcolorbox}
\usepackage{caption}
\usepackage{subcaption}
\usepackage{amsthm}
\usepackage{amssymb}
\usepackage[noabbrev,capitalize,nameinlink]{cleveref}
\usepackage{forest}
\usepackage[ruled]{algorithm2e}
\DontPrintSemicolon

\usepackage[frozencache,cachedir=.]{minted}
\AfterEndEnvironment{minted}{\vspace{-0.8cm}}

\DeclareMathOperator{\epsmin}{\varepsilon_{\text{min}}}

\DeclareMathOperator{\cX}{\mathcal{X}}
\DeclareMathOperator{\cA}{\mathcal{A}}
\DeclareMathOperator{\cB}{\mathcal{B}}

\newcommand{\indicator}[1]{\mathbbm{1}(#1)}

\newtheorem{example}{Example}
\newtheorem{theorem}{Theorem}
\newtheorem{definition}{Definition}

\newtheorem{summary}{Summary}

\newcommand{\extodo}[1]{{\color{red!70!black}\textbf{[TODO: #1]}}}

\definecolor{bg}{rgb}{0.96,0.96,0.96}

\title{\texttt{deep-significance} - Easy and Meaningful Statistical Significance Testing in the Age of Neural Networks}

  \author{Dennis Ulmer\textsuperscript{$\clubsuit$} \hspace{1em}
    Christian Hardmeier\textsuperscript{$\clubsuit$} \hspace{1em}
    Jes Frellsen\textsuperscript{$\diamondsuit$}\\
\textsuperscript{$\clubsuit$}Department of Computer Science, IT University of Copenhagen, Denmark \\
\textsuperscript{$\diamondsuit$}Department of Applied Mathematics and Computer Science, Technical University of Denmark\\
\texttt{dennis.ulmer@mailbox.org}
}

\iclrfinalcopy

\begin{document}

\maketitle

\begin{abstract}
  A lot of Machine Learning (ML) and Deep Learning (DL) research is of an empirical nature. Nevertheless, statistical significance testing (SST) is still not widely used. This endangers true progress, as seeming improvements over a baseline might be statistical flukes, leading follow-up research astray while wasting human and computational resources. Here, we provide an easy-to-use package containing different significance tests and utility functions specifically tailored towards research needs and usability.
\end{abstract}

\section{Introduction}

Deep Learning is a rapidly moving research field. Since its Cambrian explosion a decade ago, model architectures such as the transformer \citep{vaswani2017attention} have caused a paradigm shift in the field of Natural Language Processing (NLP), and most recently also in Computer Vision (CV; \citeauthor{dosovitskiy2021image}, \citeyear{dosovitskiy2021image}). Unsurprisingly, a flurry of improvements has  been proposed to enhance the original design even further. \citet{narang2021transformer} try to replicate many of these results at a large scale, finding that many of them do not improve consistently over the original model. Another very active area of DL research lies in optimizers, where Adam \citep{kingma2015adam} is often seen as the de-facto standard choice. Nonetheless, countless additions and alternatives have been put forth in this context, with equally disappointing results when benchmarked on a large scale \citep{schmidt2020descending}.

These two examples should not be considered outliers, but emblematic of a larger trend.  \emph{Statistical significance testing} is one tool that can help to identify noteworthy contributions among the noise, but remains underutilized in ML research, as exemplified by a recent meta-study in Neural Machine Translation \citep{marie2021scientific}. We give two potential reasons for these phenomena: On the one hand, significance testing still is not a standard part of the experimental workflow as in other empirical disciplines, with little external incentive given to researchers to use it. Secondly, practitioners often seem to shy away from statistical significance tests due to their (seeming) complexity, being afraid to misuse them and thereby potentially weakening the cogency of their work. This paper tries to mitigate the latter problem in order to tackle the former.

This work makes the following contributions:
\begin{itemize}
    \item We describe an assumption-less and statistically powerful significance test recently proposed in the NLP literature called \emph{Almost Stochastic Order} (ASO; \citeauthor{del2018optimal}, \citeyear{del2018optimal}; \citeauthor{dror2019deep}, \citeyear{dror2019deep}) and re-implement it alongside other, general-purpose tests in an easy-to-use open-source software package.\footnote{\url{https://github.com/Kaleidophon/deep-significance}}
    \item We include a comprehensive guide for the usage and underlying methods of the package, explaining potential limits and pitfalls of the implemented methods.
    \item We evaluate the methods against other significance test and demonstrate themin a case study.
\end{itemize}

\section{Related Work}

\paragraph{Studies of Current Trends in ML} Besides the aforementioned examples of \citet{narang2021transformer, schmidt2020descending, marie2021scientific}, other authors have noted problems with experimental standards. \citet{henderson2018deep, agarwal2021deep} analyse problems with Reinforcement Learning (RL) results. \citet{gunderson2018state} highlight problems with reproducibility and replicability in AI research. \citet{gehrmann2022repairing} comprehensively discuss issues with Natural Language Generation research, including statistical significance -- something that \citet{marie2021scientific} investigate for neural machine translation, specifically. \citet{berg2021empirical} and \citet{card2020with} investigate the limitations of $p$-values and statistical power in NLP.

\paragraph{Problems with comparing Neural Networks} Several authors have characterized problems with the direct comparisons of performance in neural network algorithms, mostly rooted in their stochastic properties. \citet{reimers2018comparing} postulate that single performance scores are insufficient to draw conclusions about performance, due to the existence of local minima with different degrees of generalization. A similar point is raised by \citet{dehghani2021benchmark}, who analyze the current state of benchmarking and criticize that comparisons of single scores necessarily yield positive results when experiments are repeated often enough. \citet{bouthillier2021accounting} identify the different sources of randomness for an algorithm and show that varying as many sources as possible between runs actually \emph{decreases} the variance of the true performance estimate. \citet{cooper2021hyperparameter} provide a formal proof that random hyperparameter optimzation can shield against contradictory conclusions about performance compared to grid search. \citet{dodge2019show} argue that performance scores should not only be seen in isolation, but also be reported in relation to the used computational budget.

\paragraph{Proposing new methodologies for Experimental Analyses} \citet{dror2019deep} propose a new statistical test to compare Deep Neural Networks, which is explained in detail in \cref{sec:aso}. \citet{dror2018hitchhiker} and \citet{azer2020not} enumerate several frequentist and Bayesian hypothesis tests, with the latter providing implementations in a software package. \citet{agarwal2021deep} provide open-source code specifically tailored towards Reinforcement Learning. \citet{wang2019bayes} derive Bayesian hypothesis tests for precision, recall and $F_1$-score. \citet{benavoli2017time} provide a tutorial for Bayesian analyses for ML as an alternative to statistical hypothesis testing, entirely.

\section{Nomenclature and Notation}

We first lay out some necessary notation and definitions to avoid confusion in the subsequent sections. In experiments, we are often interested in determining whether a new \emph{algorithm} $\mathbb{A}$ performs better than some baseline algorithm $\mathbb{B}$ on some dataset $\mathbb{D}$. We hereby utilize the following definitions:

\begin{definition}[Learning Algorithm]
    We define an algorithm $\mathbb{A}$ to be the set of predictors $\{f_\theta\}_{\theta \in \Theta}$ with a) the same parameterization $\theta \in \Theta$ and b) the same optimization procedure.
\end{definition}

\begin{definition}[Observation]
    Let us define $m: \{f_\theta\}_{\theta \in \Theta} \times \mathcal{D} \rightarrow \mathbb{R}$ to be a function measuring the performance of a predictor $f_\theta$ on some dataset $\mathbb{D} \in \mathcal{D}$ in form of a real number $s \in \mathbb{R}$, called \emph{observation} or \emph{score}. We will assume in the following that a higher number indicates a more desirable behavior. Furthermore, let $\mathbb{S}_\mathbb{A}$ denote a set of observations obtained from different instances of the algorithm $\mathbb{A}$.
\end{definition}

Ideally for Deep Neural Networks, obtaining a set of observations $\mathbb{S}_\mathbb{A}$ would ideally involve training multiple \emph{instances} of a network with the same architecture using different sets of hyperparameters and random initializations. Since the former part often becomes computationally infeasible in practice, we follow the advice of \citet{bouthillier2021accounting} and assume that it is obtained by fixing one set of hyperparameters after a prior search and varying as many other random elements as possible.



\section{Statistical Significance Testing}

Here, we only give a very brief introduction into statistical significance testing using $p$-values, and refer the reader to resources such as \citet{japkowicz2011evaluating, dror2018hitchhiker, raschka2018model, azer2020not, dror2020statistical, riezler2021validity} for a more comprehensive overview. Using the notation introduced in the previous section, we can define a one-sided test statistic $\delta(\mathbb{S}_\mathbb{A}, \mathbb{S}_\mathbb{B})$ based on the gathered observations. An example of such test statistics is for instance the difference in observation means. We then formulate the following null-hypothesis:

\begin{equation*}
    H_0: \delta(\mathbb{S}_\mathbb{A}, \mathbb{S}_\mathbb{B}) \le 0
\end{equation*}

 That means that we actually assume the opposite of our desired case, namely that $\mathbb{A}$ is not better than $\mathbb{B}$, but equally as good or worse, as indicated by the value of the test statistic. Usually, the goal becomes to reject this null hypothesis using the SST.
$p$-value testing is a frequentist method in the realm of SST. It introduces the notion of data that \emph{could have been observed} if we were to repeat our experiment again using the same conditions, which we will write with superscript $\text{rep}$ in order to distinguish them from our actually observed scores \citep{gelman2021bayesian}. We then define the $p$-value as the probability that, under the null hypothesis, the test statistic using replicated observation is larger than or equal to the \emph{observed} test statistic:

\begin{equation*}
    p(\delta(\mathbb{S}_\mathbb{A}^\text{rep}, \mathbb{S}_\mathbb{B}^\text{rep}) \ge \delta(\mathbb{S}_\mathbb{A}, \mathbb{S}_\mathbb{B})|H_0)
\end{equation*}

We can interpret this expression as follows: Assuming that $\mathbb{A}$ is not better than $\mathbb{B}$, the test assumes a corresponding distribution of statistics that $\delta$ is drawn from. So how does the observed test statistic $\delta(\mathbb{S}_\mathbb{A}, \mathbb{S}_\mathbb{B})$ fit in here? This is what the $p$-value expresses: When the probability is high, $\delta(\mathbb{S}_\mathbb{A}, \mathbb{S}_\mathbb{B})$ is in line with what we expected under the null hypothesis, so we can \emph{not} reject the null hypothesis, or in other words, we \emph{cannot} conclude $\mathbb{A}$ to be better than $\mathbb{B}$. If the probability is low, that means that the observed $\delta(\mathbb{S}, \mathbb{S}_\mathbb{B})$ is quite unlikely under the null hypothesis and that the reverse case is more likely -- i.e.\@ that it is likely larger than -- and we conclude that $\mathbb{A}$ is indeed better than $\mathbb{B}$. Note that \textbf{the $p$-value does not express whether the null hypothesis is true}. To make our decision about whether or not to reject the null hypothesis, we typically determine a threshold -- the significance level $\alpha$, often set to $0.05$ -- that the $p$-value has to fall below. However, it has been argued that a better practice involves reporting the $p$-value alongside the results without a pidgeonholing of results into significant and non-significant \citep{wasserstein2019moving}. The intuition of a $p$-value is summarized below:

\begin{tcolorbox}[colback=green!20!white,colframe=green!60!black]
    \begin{summary}[$p$-values]
        Assuming the null-hypothesis to be true, how likely is a test statistic at least as extreme as observed?
    \end{summary}
\end{tcolorbox}

\subsection{Almost Stochastic Order}\label{sec:aso}

Deep neural networks are known to be highly non-linear models \citep{li2018visualizing}, having their performance depend to a large extent on the choice of hyperparameters, random seeds and other (stochastic) factors \citep{bouthillier2021accounting}. This makes comparisons between algorithms more difficult, as illustrated by the motivating example below by \citet{dror2019deep}:

\begin{tcolorbox}[colback=blue!10,colframe=blue!40!white]
    \begin{minipage}{0.6\textwidth}
        \begin{example}[Part-of-Speech tagging]
            Consider the results for a PoS-tagging task (assigning categories like verb or noun to words) given in the table on the right, taken over 3898 and 1822 observations using different hyperparameter configurations and random seeds, respectively. Using the Adam optimizer \citep{kingma2015adam} for a model gives a higher average word-level accuracy than using RMSprop \citep{tieleman2012lecture}, however the median score favors the latter. Furthermore, the minimum result across a few runs favor Adam, but the maximum score is higher for RMSprop. So, which algorithm do we consider to be \emph{better}?
        \end{example}
    \end{minipage}
    \begin{wraptable}[10]{r}{4.3cm}
        \vspace{-4.5cm}
        \setlength{\tabcolsep}{6pt}
        \renewcommand{\arraystretch}{1.4}
        \begin{tabular}{@{}lrr@{}}
            \toprule
            & Adam & RMSprop  \\
            \midrule
            Mean & 0.9224 & 0.9190 \\
            Std. dev. & 0.0604 & 0.0920 \\
            Median & 0.9319 & 0.9349 \\
            Min. & 0.1746 & 0.1420 \\
            Max. & 0.9556 & 0.9573 \\
            \bottomrule
        \end{tabular}
    \end{wraptable}
\end{tcolorbox}

Therefore, \citet{dror2019deep} propose \emph{Almost Stochastic Order} for Deep Learning models based on the work by \citet{del2018optimal}. It is based on a relaxation of the concept of \emph{stochastic order} by \citet{lehmann1955ordered}: A random variable $\cX_{\cA}$ is defined to be \emph{stochastically larger} than $\cX_{\cB}$ (denoted $\cX_{\cA} \succeq \cX_{\cB}$) if $\forall x: F(x) \le G(x)$, where $F$ and $G$ denote the cumulative distribution functions (CMF) of the two random variables. The CDF is  defined as $F(t) = p(\cX \le t)$, while the \emph{empirical} CDF given a sample $\{x_1, \ldots, x_n\}$ is defined as

\begin{equation*}
    F_n(t) = \frac{1}{n}\sum_{i=1}^n \indicator{x_i \le t}
\end{equation*}

\begin{figure}[bt]
    \centering
    \begin{subfigure}[t]{0.425\textwidth}
        \includegraphics[width=\textwidth]{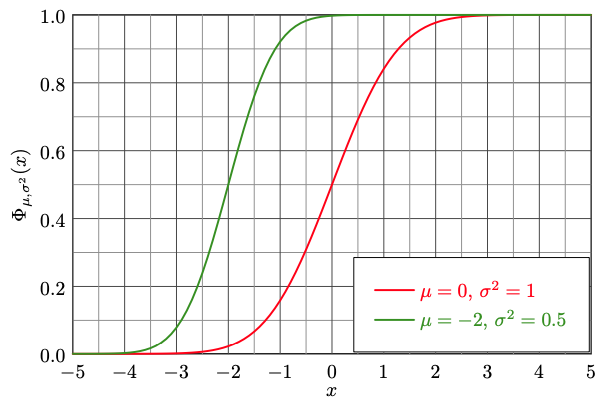}
        \caption{Stochastic order with red $\succeq$ green.}
        \label{subfig:so}
    \end{subfigure}
    \hspace{0.2cm}
    \begin{subfigure}[t]{0.425\textwidth}
        \includegraphics[width=\textwidth]{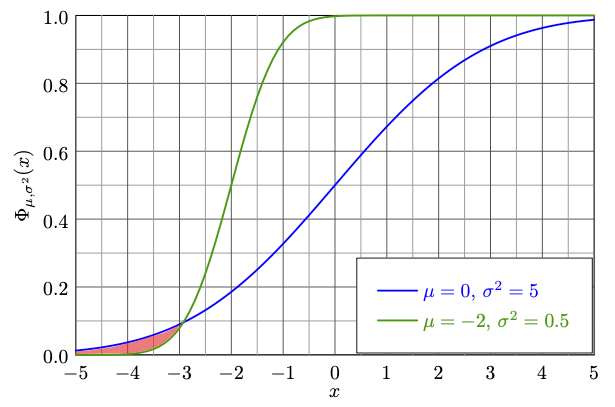}
        \caption{Almost stochastic order, with blue $\succsim$ green.}
        \label{subfig:aso}
    \end{subfigure}
    \caption{Examples for stochastic order (a) and almost stochastic order (b), illustrated using the CDFs of two normal random variables. Because stochastic order is too strict to be practical, almost stochastic order allows for some degree of violation of the order (red area in (b)).}\label{fig:aso-so}
\end{figure}

with $\indicator{\cdot}$ being the indicator function. In practice, since we do not know the real score distributions $p(\cX_{\cA})$ and $p(\cX_{\cB})$, we cannot use the precise CDFs in subsequent calculations, and we rely on the empirical CDFs $F_n$ and $G_m$. A case of stochastic order is illustrated in \cref{subfig:so}, using the CDFs of two normal distributions. However, in cases such \cref{subfig:aso} we would still like to declare one of the algorithms superior, even though the stochastic order of the underlying CDFs is partially violated. Several ways to quantify the violation of stochastic dominance exist \citep{alvarez2017models,del2018some}, but here we elaborate on the optimal transport approach by \citet{del2018optimal}. They propose a the following expression quantifying the distance of each random variables from being stochastically larger than the other:

\begin{equation}\label{eq:epsilon-dist}
    \varepsilon_{W_2}(F, G) = \frac{\int_{\mathbb{V}_{\cX}} (F^{-1}(t) - G^{-1}(t))^2 dt}{(W_2(F, G))^2}
\end{equation}

with the \emph{violation ratio} $\varepsilon_{W_2}(F, G) \in [0, 1]$ and a \emph{violation set} $\mathbb{V}_{\cX} = \big\{t \in (0, 1): F^{-1}(t) < G^{-1}(t) \big\}$, i.e.\@ where the stochastic order is being violated. \cref{eq:epsilon-dist} contains the following components: Firstly, the quantile functions $F^{-1}(t)$ and $G^{-1}(t)$ associated with the correspondng CDFs:

\begin{equation*}
    F^{-1}(t) = \inf \big\{x: t \le F(x)\big\}, \quad t \in (0, 1)
\end{equation*}

The quantile functions allow us to define stochastic order via $X \succeq Y \iff \forall t \in (0, 1): F^{-1}(t) \ge G^{-1}(t)$. Secondly, the univariate $l_2$-Wasserstein distance:

\begin{equation}\label{eq:wasserstein}
    W_2(F, G) = \sqrt{\int_0^1 \big(F^{-1}(t) -  G^{-1}(t)\big)^2 dt}
\end{equation}

Finally, \citet{del2018optimal, dror2019deep} define a hypothesis test based on this quantity by formulating the following hypotheses:

\begin{equation*}\begin{aligned}
    H_0:\ & \varepsilon_{W_2}(F, G) \ge \tau\\
    H_1:\ & \varepsilon_{W_2}(F, G) < \tau \\
\end{aligned}\end{equation*}

for a pre-defined threshold $\tau > 0$, for instance  $0.5$ or lower (see discussion in \cref{app:aso-type1-error} about the choice of threshold). Further, \citet{alvarez2017models, dror2019deep} produce a frequentist upper bound to this quantity, defining the minimal $\varepsilon_{W_2}$ for which we can reject the null hypothesis with a confidence of $1 - \alpha$ as

\begin{equation}\label{eq:epsmin}
    \epsmin(F_n, G_m, \alpha) = \varepsilon_{W_2}(F_n, G_m) - \sqrt{\frac{n+m}{nm}}\hat{\sigma}_{n, m}\Phi^{-1}(\alpha)
\end{equation}

The variance term $\hat{\sigma}_{n, m}$ is estimated using bootstrapping, with $F_n^*$ and $G_m^*$ denoting empirical CDFs based on sets of scores resampled from original sets of model scores, similar to re-sampling procedure in other tests like the bootstrap \citep{efron1994introduction} or permutation-randomization test \citep{noreen1989computer}:

\begin{equation}\label{eq:aso-var}
    \hat{\sigma}_{n, m}^2 = \text{Var}\bigg[\sqrt{\frac{mn}{n + m}}\big(\varepsilon_{W_2}(F_n^*, G_m^*) - \varepsilon_{W_2}(F_n, G_m)\big) \bigg]
\end{equation}

Thus, if $\epsmin(F_n, G_m, \alpha) < \tau$, we can reject the null hypothesis and claim that algorithm $\mathbb{A}$ is better than $\mathbb{B}$, with a growing discrepancy in performance the smaller the value becomes. This enables us to pose the following kind of hypotheses:

\begin{tcolorbox}[colback=green!20!white,colframe=green!60!black]
    \begin{summary}[Almost Stochastic Order]
        Given the observed scores and a confidence of $1 - \alpha$, what is the expected upper bound to the violation ratio of algorithm's $\mathbb{A}$ empirical CDF over $\mathbb{B}$'s?
    \end{summary}
\end{tcolorbox}

\section{Experimental Comparison with other Tests}\label{sec:experiments}

\begin{wrapfigure}[16]{R}{0.4\textwidth}
    \centering
    \includegraphics[width=0.4\textwidth]{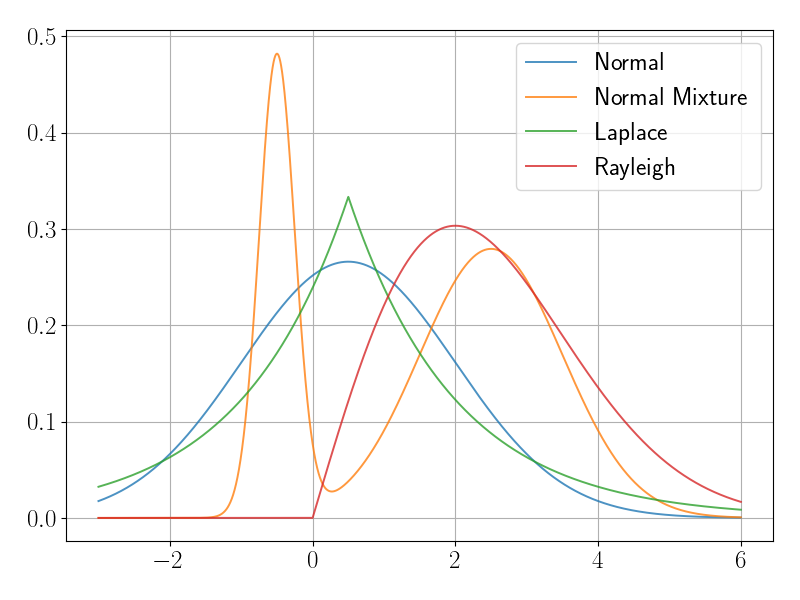}
    \caption{Plot of distributions used for Type I and Type II error tests.}\label{fig:distributions}
\end{wrapfigure}

\begin{figure}[h]
    \centering
    \begin{subfigure}[t]{0.485\textwidth}
        \includegraphics[width=\textwidth]{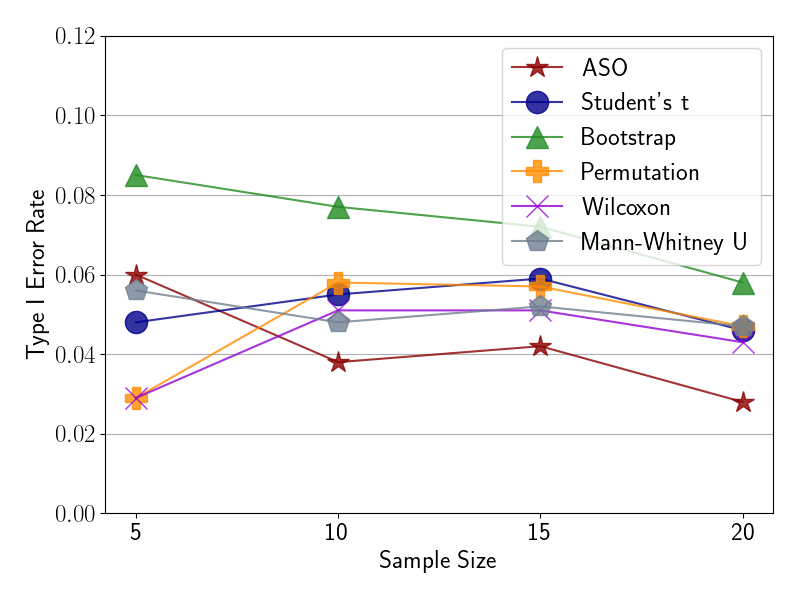}
        \caption{Rates for normal samples.}
        \label{subfig:type1_size_normal}
    \end{subfigure}%
    \hfill
    \begin{subfigure}[t]{0.485\textwidth}
        \includegraphics[width=\textwidth]{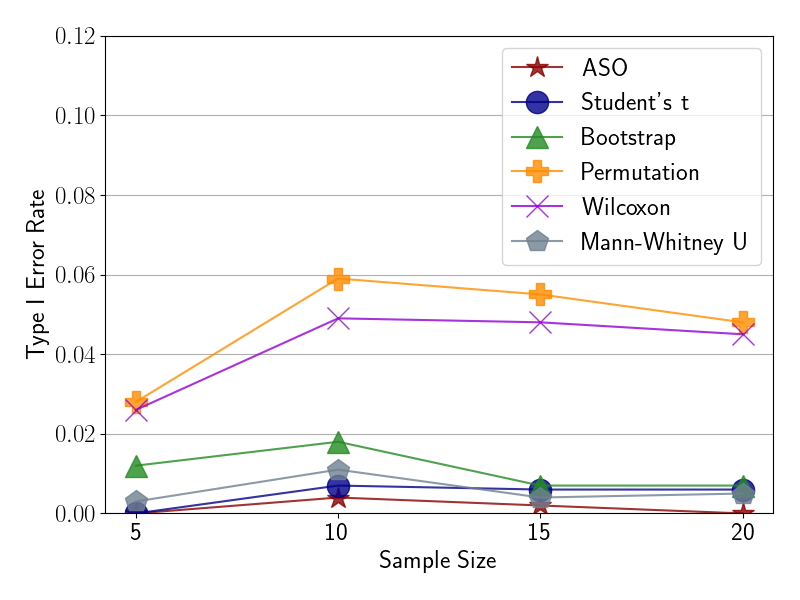}
        \caption{Rates for normal mixture samples.}
        \label{subfig:type1_size_normal_mixture}
    \end{subfigure}%

    \begin{subfigure}[t]{0.485\textwidth}
        \includegraphics[width=\textwidth]{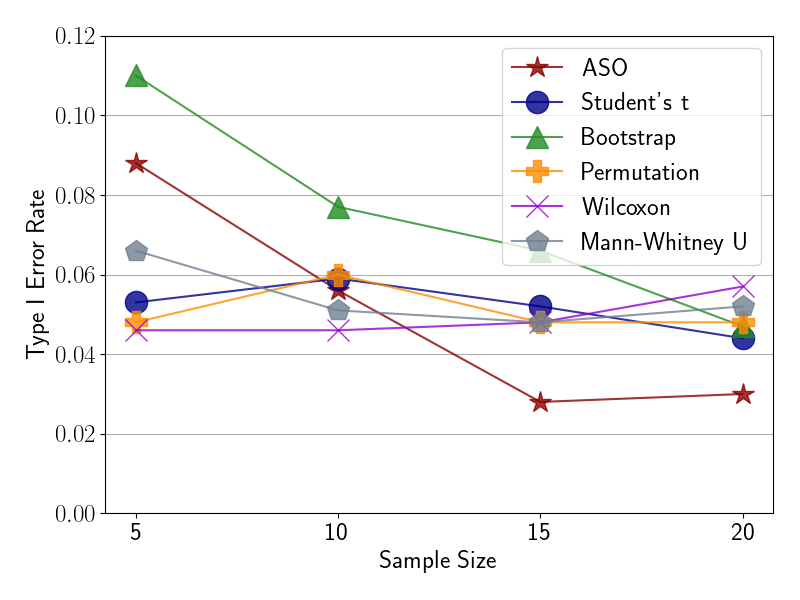}
        \caption{Rates for Laplace samples.}
        \label{subfig:type1_size_laplace}
    \end{subfigure}%
    \hfill
    \begin{subfigure}[t]{0.487\textwidth}
        \includegraphics[width=\textwidth]{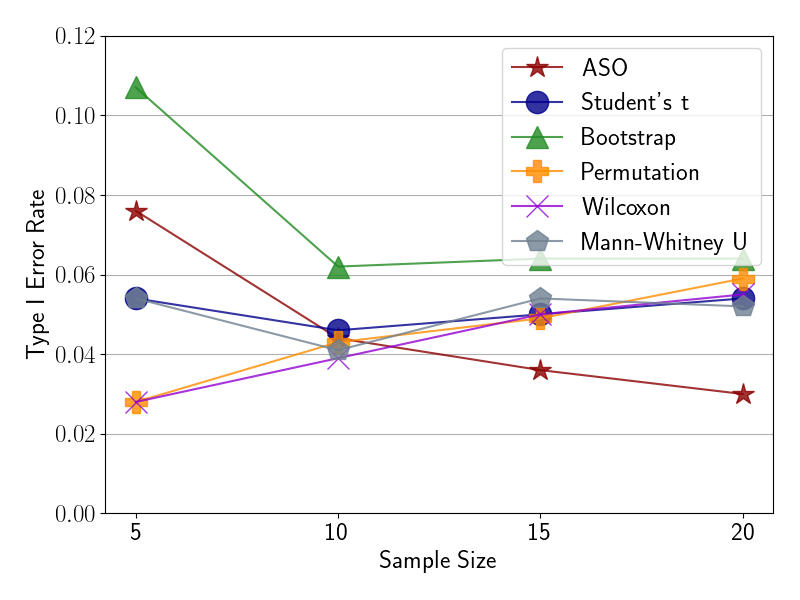}
        \caption{Rates for Rayleigh samples.}
        \label{subfig:type1_size_rayleigh}
    \end{subfigure}
    \caption{Comparing type I error rates for different tests and distributions as a function of sample size. Decisions are made using a confidence threshold of $\alpha = 0.05$ and $\tau = 0.2$ for $\varepsilon_\text{min}$.}\label{fig:aso-tests}
\end{figure}

We compare ASO to established significance tests such as the Student's t, the bootstrap \citep{efron1994introduction}, permutation-randomization test \citep{noreen1989computer} along with the Wilcoxon signed-rank \citep{wilcoxon1992individual} and Mann-Whitney U test \citep{mann1947test} on different types of distributions, which are plotted in \cref{fig:distributions}. We plot the Type I error rate per $500$ simulations for ASO and $1000$ simulations for the other tests as a function of sample size in \cref{fig:aso-tests}, where we sample both sets of observation from the same distribution. For \cref{subfig:type1_size_normal}, we sample from $\mathcal{N}(0, 1.5^2)$ and try a bimodal normal mixture in \cref{subfig:type1_size_normal_mixture} (using the same parameter for the second component, and $\mathcal{N}(-0.5, 0.25^2)$ with mixture weights $\pi_1=0.75$ and $\pi_2=0.25$). To also test the behavior of tests on non-normal distributions, we also sample from a $\text{Laplace}(0, 1.5^2)$ distribution in \cref{subfig:type1_size_laplace}, which posseses a different behavior around the main, as well as the Rayleigh distribution with $\text{Rayleigh}(1)$ in \cref{subfig:type1_size_rayleigh}, which has a heavy tail.

We can see that ASO performs either en par or better than other tests in all scenarios, achieving \emph{lower} error rates the more samples are available, while other tests score around the expected type I error of $5 \%$. In \cref{app:experiments}, Type II error experiments reveal that the test produces comparatively higher error rates for ASO, though. This can be explained by the fact that we use the upper bound $\epsmin$ instead of $\varepsilon_{W_2}$ to evaluate the null hypothesis, which makes the test act more conservatively. We also find in \cref{app:experiments} that a decision threshold of $\tau = 0.2$ strikes an acceptable balance between Type I and II error rates across different scenarios. Overall, we argue that a lower Type I error is more advantageous in the context of empirical research, and that a \emph{decreasing} error rate w.r.t.\@ higher sample sizes constitutes an appealing property when used on arbitrary distributions.

In these experiments, the score distributions were determined \emph{a priori} in order to create rigid experimental conditions. Naturally, a practitioner would not know these distribution in a usual setting, which is why we illustrate the usage of the package the next section.

\section{Package Contents \& Examples}\label{sec:examples}

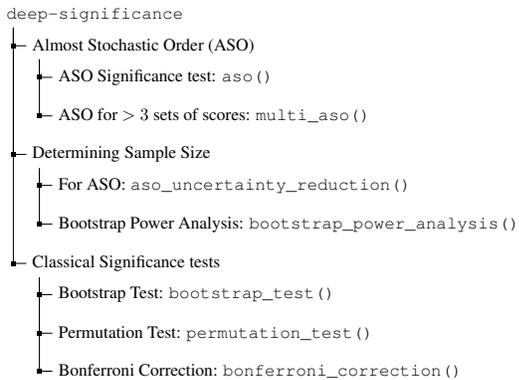
\begin{wrapfigure}[16]{R}{0.5\textwidth}
    \vspace{-0.5cm}
    \centering
    \resizebox{0.5\textwidth}{!}{%
    \begin{forest}
      for tree={
        grow'=0,
        child anchor=west,
        parent anchor=south,
        anchor=west,
        calign=first,
        edge path={
          \noexpand\path [draw, \forestoption{edge}]
          (!u.south west) +(7.5pt,0) |- node[fill,inner sep=1.25pt] {} (.child anchor)\forestoption{edge label};
        },
        before typesetting nodes={
          if n=1
            {insert before={[,phantom]}}
            {}
        },
        fit=band,
        before computing xy={l=15pt},
      }
      [\texttt{deep-significance}
          [Almost Stochastic Order (ASO)
            [ASO Significance test: \texttt{aso()}]
            [ASO for $>3$ sets of scores: \texttt{multi\_aso()}]
          ]
          [Determining Sample Size
            [For ASO: \texttt{aso\_uncertainty\_reduction()}]
            [Bootstrap Power Analysis: \texttt{bootstrap\_power\_analysis()}]
          ]
          [Classical Significance tests
            [Bootstrap Test: \texttt{bootstrap\_test()}]
            [Permutation Test: \texttt{permutation\_test()}]
            [Bonferroni Correction: \texttt{bonferroni\_correction()}]
          ]
      ]
    \end{forest}%
    }
    \caption{Overview over package contents of \texttt{deep-significance v1.2.5}.}
    \label{fig:contents}
\end{wrapfigure}

\cref{fig:contents} lists the three main groups contents in the package. For one, the package implements some common significance tests, such as the permutation-randomization test \citep{noreen1989computer} and the bootstrap test \citep{efron1994introduction}, as well as function to perform the Bonferroni correction \citep{bonferroni1936teoria} for multiple comparisons. These tests where chosen because they do not come with any assumptions about the distribution on samples -- which also implies low statistical power.

Another part of the package is dedicated to determining the right sample size -- \texttt{aso\_uncertainty\_reduction()} determine the factor by which the uncertainty about the estimate of $\varepsilon_{W_2}(F, G)$ is reduced by increasing the size of either sample of scores. Another, more general approach is Bootstrap Power Analysis \citep{yuan2003bootstrap, henderson2018deep}: Each score in the sample is given an equal lift by a constant factor, creating a second, artificial sample. Then, a significance test is run repeatedly on bootstrapped versions of both samples, the percentage of resulting $p$-values under a given threshold is recorded. Ideally, this should result in large number of significant differences induced by the lift -- if not, this can be indicate a too high of a variance in the original set of scores. Lastly, the package implements the ASO test from \cref{sec:aso}, with more details described in \cref{app:implementation-details}. 

\paragraph{Quality-of-Life features} To increase ease of use, the package comes with the following features: Scores can be supplied in the most common data types, including Python lists, NumPy \citep{harris2020array} and JAX arrays
\citep{jax2018github}, as well as PyTorch \citep{paszke2017automatic} and Tensorflow tensors \citep{tensorflow2015-whitepaper}. To decrease the waiting time for results, the number of processes can be increases using the \texttt{num\_jobs} argument. Furthermore, for comparing more than two samples at once, \texttt{multi\_aso()} can be used, which outputs the results in a tabular structure\footnote{Including the option of returning the results in a pandas \texttt{DataFrame} \citep{mckinney-proc-scipy-2010,reback2020pandas}, which itself can easily be converted into a \LaTeX\ table.} and automatically applies the Bonferroni correction by default. All stochastic functions support seeding for replicability.

\paragraph{Choice of Test} All packaged tests come with weak or no assumption about the score distribution. If the distribution is known, an appropriate parametric test should always be preferred. The bootstrap and permutation test can be used if $p$-values are desired, with the former possessing a higher statistical power. In cases with unusual or unknown score distributions, we recommend ASO, since the only assumptions it makes are that the true CDFs $F$ and $G$ have bounded convex support and their underlying PDFs to have finite second order moments.

\subsection{Case study -- Deep Q-Learning}\label{sec:case-study}

In order to demonstrate the intended use of the package, we showcase its use in a case study, based on the simple Reinforcement Learning described below. Here, reward distributions are usually not normal, and therefore provide an ideal testbed. All used code is available in the repository.\footnote{\url{https://github.com/Kaleidophon/deep-significance/blob/paper/paper/deep-significance}}

\begin{tcolorbox}[colback=blue!10,colframe=blue!40!white]

    \begin{minipage}{0.6\textwidth}
        \begin{example}[Cart Pole Problem; \citeauthor{barto1983neuronlike}, \citeyear{barto1983neuronlike}]
            In the cart pole problem, a reinforcement learning agent's goal is to balance the eponymous pole on a cart that is only allowed to move in a horizontal direction. When the pole falls over, the episode ends. A reward of $+1$ is awarded for every time step during which the pole remains upright.
        \end{example}
    \end{minipage}
    \begin{minipage}{0.38\textwidth}
    \begin{center}
        \hfill
        \includegraphics[width=0.85\textwidth]{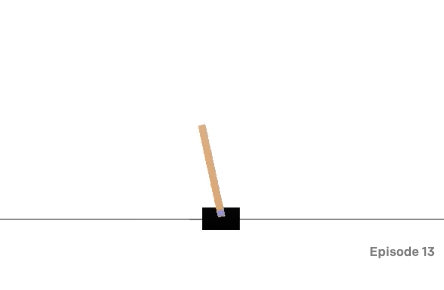}
        \\ \hspace{0.3cm} {\footnotesize Cart Pole environment by \citet{openai2016brockman}}.
       \end{center}
    \end{minipage}
\end{tcolorbox}

We tackle the problem above with a classic Deep Reinforcement Learning approach, namely \emph{Deep Q-Learning} \citep{mnih2015human}. Deep Q-Learning tries to approximate the optimal action-value function defined as

\begin{equation*}
    Q^*(s, a) = \max_\pi \mathbb{E}\big[ r_t + \gamma r_{t+1} + \gamma^2 r_{t+2} + \ldots \big| s_t = s, a_t = a, \pi \big]
\end{equation*}

The definition above reads as follow: The optimal action-value function is the policy $\pi$ that maximizes the future reward $r_t$ at a state $s_t$ by performing an action $a_t$, with subsequent rewards being increasingly discounted by a factor $\gamma$. The model weights are updated using the following $l_2$ loss:

\begin{equation*}
    \mathcal{L}(\theta) = \mathbb{E}_{(s, a, r, s^\prime) \sim U(\text{Buffer})}\bigg[\Big(r + \max_{a^\prime} Q(s^\prime, a^\prime; \theta^\text{target}) - Q(s, a; \theta)\Big)^2\bigg]
\end{equation*}

Two aspects of this loss function are especially noteworthy: First of all, since we do not know the true value of the $Q$-function in most cases, the predicted value $Q(s, a; \theta)$ is compared against the reward plus outcome of the greedy action chosen by a \emph{target} network: To avoid having to ``hit a moving target'' \citep{van2018deep}, the target network is only updated every couple of training steps by copying the main networks parameters. Secondly, the state, action and reward used to compute the loss are not the ones just observed by the model, but instead are uniformly sampled from a \emph{replay buffer}, a sort of memory that past experiences gets added to during training.

In our example, we investigate how the update frequency of the target network affects the mean of the rewards obtained during training. First, we train five models with an update frequency of $10$ and $20$ steps each, and store the results. But how do we know that we collected enough scores to make a meaningful comparison? The package supplies two functions for this purpose, the first of which is an implementation of bootstrap power analysis \citep{yuan2003bootstrap} shown below:

\begin{minted}[breaklines, breakafter=d, fontfamily=tt, fontsize=\footnotesize, style=sas, fontshape=bold, bgcolor=bg, numbersep=2pt, python3=True, autogobble=true]{python}
    from deepsig import bootstrap_power_analysis

    bootstrap_power_analysis(reward_dist_freq_10, num_jobs=4) # Results for 10 steps, gives 0.6594
    bootstrap_power_analysis(reward_dist_freq_20, num_jobs=4) # Results for 20 steps, gives 0.5616
\end{minted}

These scores have a direct statistical interpretation, since they signify the \emph{statistical power}. The higher the statistical power, the lower the probability of a Type II error or false negative. A common rule of thumb is to thrive for a power of around $0.8$, and we might therefore want to collect more samples here. For instance, we could decide to collect $10$ or $15$ samples in total. In case we are using the ASO test, the second function can help with this decision:

\begin{minted}[breaklines, breakafter=d, fontfamily=tt, fontsize=\footnotesize, style=sas, bgcolor=bg, autogobble=true]{python}
    from deepsig import aso_uncertainty_reduction

    aso_uncertainty_reduction(m_old=5, n_old=5, m_new=10, n_new=10)  # 1.414
    aso_uncertainty_reduction(m_old=5, n_old=5, m_new=15, n_new=15)  # 1.732
\end{minted}

Since ASO only computes the ``true'' $\varepsilon_\text{min}$ value in the limit of infinitely large samples, the estimate obtained using bootstrapping has some inherent variance, which can be reduced by adding more scores to the sample. The function above computes the factor by which the uncertainty in the test result is being reduced. We can thus read the above as adding five more samples reducing the uncertainty by a factor of $1.41$, while adding ten more sample only reduces it by $1.73$. To strike a compromise with our computational budget, we decide to only add five more samples each and run two of the implemented tests:

\begin{minted}[breaklines, breakafter=d, fontfamily=tt, fontsize=\footnotesize, style=sas, bgcolor=bg, autogobble=true]{python}
    from deepsig import aso, bootstrap_test

    aso(rewards_freq_10, rewards_freq_20, num_jobs=4)  # 0.02
    bootstrap_test(rewards_freq_10, rewards_freq_20, num_jobs=4)  # 0.005
\end{minted}

\begin{wrapfigure}[8]{R}{0.4\textwidth}
    \vspace{-0.5cm}
    \centering
    \includegraphics[width=0.4\textwidth]{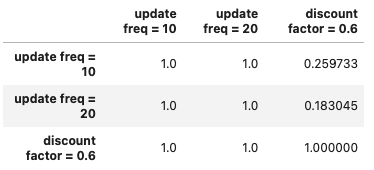}
    \caption{Results of demo.}\label{fig:demo-results}
\end{wrapfigure}

These results suggest that an update frequency of $10$ works is superior. Lastly, we can easily facilitate comparisons between multiple models using the \texttt{multi\_aso()} function below. By supplying scores in dictionary form and specifying \texttt{return\_df=True}, the results will be returned in an easily readable \texttt{pandas\ DataFrame}.  Furthermore, the Bonferroni correction \citep{bonferroni1936teoria} is being applied automatically to avoid the multiple comparisons problem.

\begin{minted}[breaklines, breakafter=d, fontfamily=tt, fontsize=\footnotesize, style=sas, bgcolor=bg, autogobble=true]{python}
    res_df = multi_aso(
        {
            "update freq = 10": reward_dist_freq_10,
            "update freq = 20": reward_dist_freq_20,
            "discount factor = 0.6": reward_dist_discount_06
        },
        num_jobs=4, return_df=True
    )
\end{minted}

Overall, the results in the demonstration shown in \cref{fig:demo-results} that an update frequency of ten steps performed best, and that using a lower discount factor produced worse results.

\section{Discussion}\label{sec:discussion}

The previous sections have demonstrated the advantages of the implemented tests in an neural network setting. Furthermore, the package implements other useful functions that can be used to evaluate experimental settings. Nevertheless, using these techniques in practice comes with limitations as well, which the end user should be aware of.

The first line of limits comes with ASO itself. Multiple steps of the procedure require different kinds of approximations or properties that are only guaranteed to hold in the infinite-sample limit , e.g.\@ \cref{eq:aso-var}. Furthermore, significance tests in general are known to sometimes provide unreliable results with small \citep{reimers2018comparing} or very large sample sizes \citep{lin2013research}, are prone to misinterpretation \citep{gibson2021role, greenland2016statistical}, and encourages binary significant / non-significant thinking \citep{wasserstein2019moving, azer2020not}.
An attractive alternative to statistical hypothesis testing therefore comes in the form of Bayesian analysis \citep{kruschke2013bayesian, benavoli2017time, gelman2021bayesian}, where the user draws conclusions from posterior distributions over quantities of interest. A potential drawback of this methodology is that it often comes at the cost of having to use Markov Chain Monte Carlo methods, and requiring experience from the user with checking convergence and defining appropriate models and model priors.



\section{Conclusion}

This work has presented an open-source software package implementing several useful tools to evaluate experimental results for Deep Neural Networks, including the ASO test by \citet{del2018optimal, dror2019deep}. We demonstrated their usefulness in a case study and discussed potential shortcomings and pitfalls. We see this package as a valuable contribution to improve experimental rigour in Machine Learning, while maintaining accessibility and ease of use. Future work could for instance derive more robust estimations of the violation ratio for small sample sizes, or come up with reliable and general Bayesian tests to evaluate experiments.

\section*{Acknowledgements}

We would like to express gratitude to Rotem Dror for supplying the plots in Figure \ref{fig:aso-so}, as well as answering questions and providing feedback to the implementation and documentation of \texttt{deep-significance}. We further thank Carlos Matr\'{a}n for answering some questions about computing the violation of stochastic order. Steady feedback to improve the usability was also supplied by members of the NLPNorth group at the IT University of Copenhagen, with special thanks going to Mike Zhang.

\bibliography{references}
\bibliographystyle{iclr2022_conference}


\appendix

\section{Implementation Details}\label{app:implementation-details}

\begin{algorithm}[h]
 \KwData{Sets of observations $\mathbb{S}_\mathbb{A}$ and $\mathbb{S}_\mathbb{B}$, integration interval $\Delta_t$, number of bootstrap iterations $B$, desired confidence level $1 - \alpha$.}
 \KwResult{Upper bound to violation ration $\epsmin$.}
 \;
 $\varepsilon_{\mathcal{W}_2}(F_n, G_m)$ = \texttt{compute\_violation\_ratio}($\mathbb{S}_\mathbb{A}$, $\mathbb{S}_\mathbb{A}$, $\Delta_t$)\;
  \;
  \tcp{Bootstrapping}
 \For{$i \in\ 0, \ldots,B$}{
  $\mathbb{S}_\mathbb{A}^* =$ \texttt{bootstrap\_sample}($\mathbb{S}_\mathbb{A}$)\;
  $\mathbb{S}_\mathbb{B}^* =$ \texttt{bootstrap\_sample}($\mathbb{S}_\mathbb{B}$)\;
   \;
   \tcp{Store value below in list}
    $\varepsilon^*_{\mathcal{W}_2}(F_n, G_m)$ = \texttt{compute\_violation\_ratio}($\mathbb{S}_\mathbb{A}^*$, $\mathbb{S}_\mathbb{A}^*$, $\Delta_t$)\;
 }
 \;
 \tcp{Compute value below based on variance of all the $\varepsilon_{\mathcal{W}_2}^*$ in list}
 $\hat{\sigma}_{n, m}^2 = \text{Var}\bigg[\sqrt{\frac{mn}{n + m}}\big(\varepsilon_{W_2}(F_n^*, G_m^*) - \varepsilon_{W_2}(F_n, G_m)\big) \bigg]$\;
 $\epsmin(F_n, G_m, \alpha) = \varepsilon_{W_2}(F_n, G_m) - \sqrt{\frac{n+m}{nm}}\hat{\sigma}_{n, m}\Phi^{-1}(\alpha)$\;

 \caption{Almost Stochastic Order (ASO) Significance Test}\label{algo:aso}
\end{algorithm}

The full algorithm to compute the $\epsmin$ score is given in Algorithm \ref{algo:aso}, and will now be explained in full detail. We show how the violation ratio in \cref{eq:epsilon-dist} can be compute in Python:

\begin{minted}[breaklines, breakafter=d, fontfamily=tt, fontsize=\footnotesize, style=sas, bgcolor=bg, autogobble=true]{python}
def compute_violation_ratio(scores_a: np.array, scores_b: np.array, dt: float) -> float:
    quantile_func_a = get_quantile_function(scores_a)
    quantile_func_b = get_quantile_function(scores_b)

    t = np.arange(dt, 1, dt)  # Points we integrate over
    f = quantile_func_a(t)  # F-1(t)
    g = quantile_func_b(t)  # G-1(t)
    diff = g - f
    squared_wasserstein_dist = np.sum(diff ** 2 * dt)

    # Now only consider points where stochastic order is being violated and set the rest to 0
    diff[f >= g] = 0
    int_violation_set = np.sum(diff[1:] ** 2 * dt)  # Ignore t = 0 since t in (0, 1)

    violation_ratio = int_violation_set / squared_wasserstein_dist

    return violation_ratio
\end{minted}

We can see that the integration over the violation set $\mathbb{V}_X$ in \cref{eq:epsilon-dist} is being performed by masking out values for which the stochastic order is honored (i.e.\@ where $F_n^{-1}(t) \ge G_n^{-1}(t)$. Computing the violation ratio involves building the empirical inverse cumulative distribution function or empircial quantile function, the same method as in \citet{dror2019deep} is used, with the corresponding Python code given below:

\begin{minted}[breaklines, breakafter=d, fontfamily=tt, fontsize=\footnotesize, style=sas, bgcolor=bg, autogobble=true]{python}
def get_quantile_function(scores: np.array) -> Callable:
    def _quantile_function(p: float) -> float:
        cdf = np.sort(scores)
        num = len(scores)
        index = int(np.ceil(num * p))

        return cdf[np.clip(index - 1, 0, num - 1))]

    return np.vectorize(_quantile_function)
\end{minted}

This function is also used inside the bootstrap sampling procedure, the last missing part of the implementation. We again follow the implementation by \citet{dror2019deep} and employ the inverse transform sampling procedure, in which we draw $p \sim \mathcal{U}[0, 1]$ and run it through a quantile function to create a sample.

\section{Experimental Appendix}\label{app:experiments}

\subsection{Additional Error Rate Experiments}\label{app:aso-type1-error}

\begin{figure}[h]
    \centering
    \begin{subfigure}[t]{0.485\textwidth}
        \includegraphics[width=\textwidth]{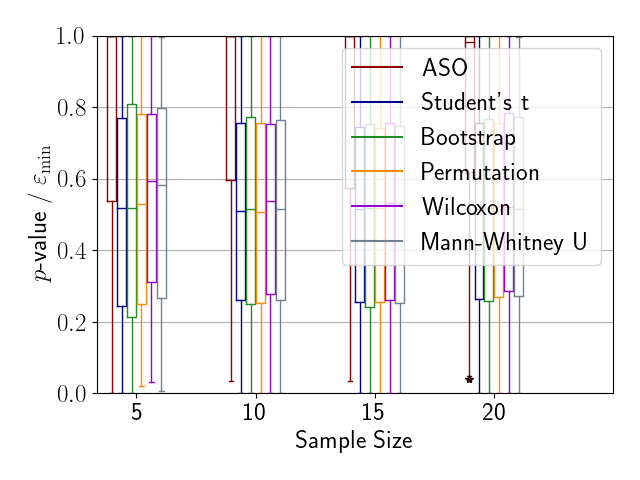}
        \caption{Dists. for normal samples.}
        \label{subfig:type1_size_normal_dists}
    \end{subfigure}%
    \hfill
    \begin{subfigure}[t]{0.485\textwidth}
        \includegraphics[width=\textwidth]{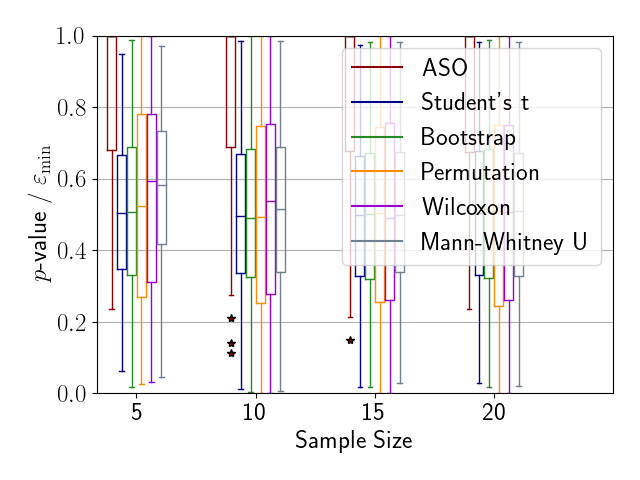}
        \caption{Dists. for normal mixture samples.}
        \label{subfig:type1_size_normal_mixture_dists}
    \end{subfigure}%

    \begin{subfigure}[t]{0.485\textwidth}
        \includegraphics[width=\textwidth]{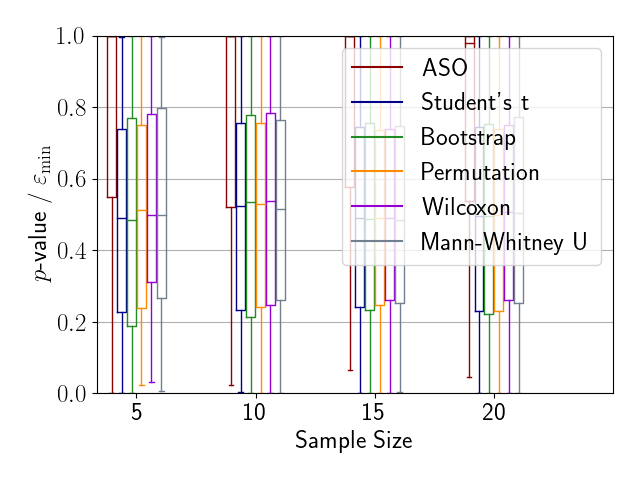}
        \caption{Dists. for Laplace samples.}
        \label{subfig:type1_size_laplace_dists}
    \end{subfigure}%
    \hfill
    \begin{subfigure}[t]{0.485\textwidth}
        \includegraphics[width=\textwidth]{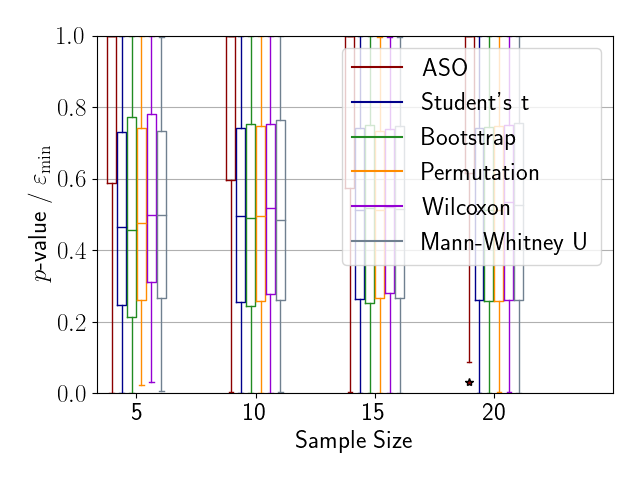}
        \caption{Dists. for Rayleigh samples.}
        \label{subfig:type1_size_rayleigh_dists}
    \end{subfigure}
    \caption{Comparing test score distributions for different tests and distributions as a function of sample size.}\label{fig:aso-tests-dists}
\end{figure}

We use this section to further shed light on the results in \cref{fig:aso-tests}.

\paragraph{Test score distributions} Instead of showing the Type I error rates based on thresholded test results, we instead plot the distributions over test scores in \cref{fig:aso-tests-dists}. We can observe that the lower ends of the interquartile range of $\epsmin$ distributions are either the same or higher than the ones for $p$-values (they do not need to be centered around $0.5$ since $\epsmin$ is an upper bound to $\varepsilon_{W_2}$), explaining the lower Type I error rate.

\begin{figure}[h]
    \centering
    \begin{subfigure}[t]{0.475\textwidth}
        \includegraphics[width=0.95\textwidth]{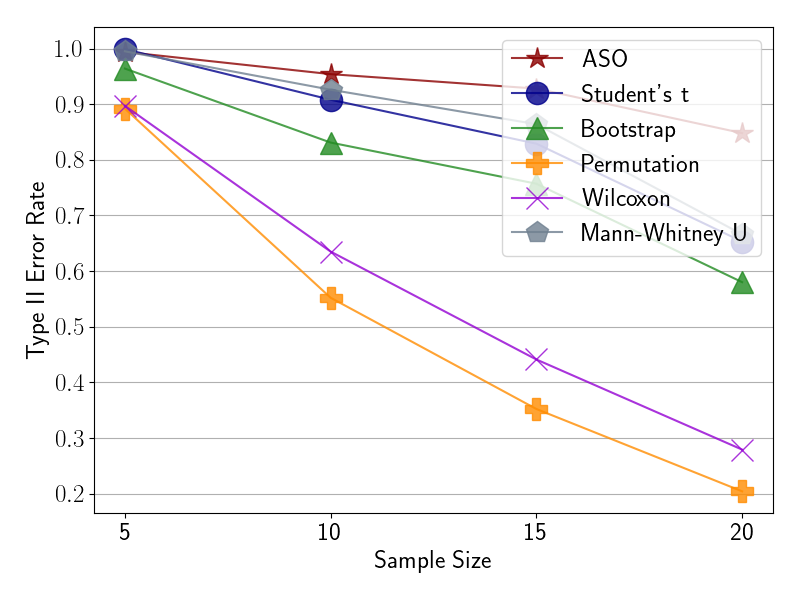}
        \caption{Type II error as a function of sample size.}
        \label{subfig:type2_normal_rates}
    \end{subfigure}%
    \hfill
    \begin{subfigure}[t]{0.475\textwidth}
        \includegraphics[width=0.95\textwidth]{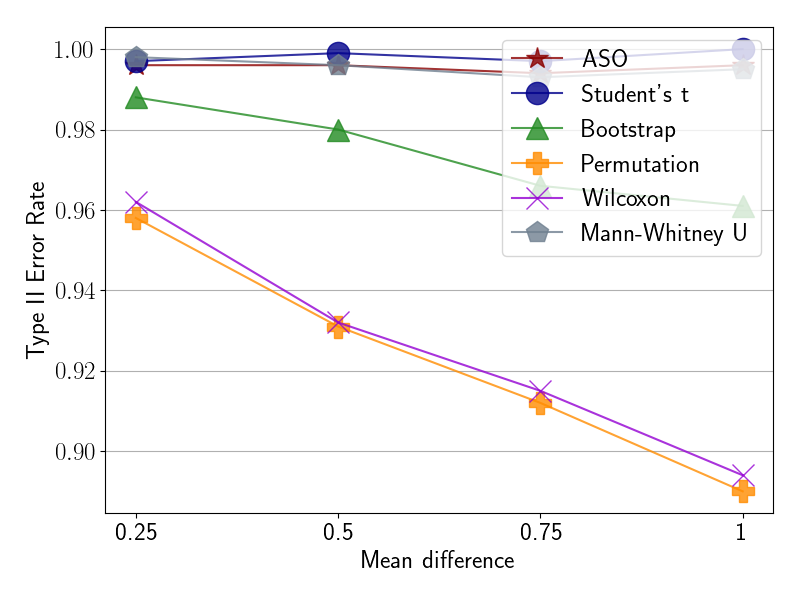}
        \caption{Type II error rate as a function of mean difference.}
        \label{subfig:type2_normal_mean_rate}
    \end{subfigure}%

    \begin{subfigure}[t]{0.475\textwidth}
        \includegraphics[width=0.95\textwidth]{img/normal_mixture/type2_rates.png}
        \caption{Type II error as a function of sample size.}
        \label{subfig:type2_normal_mixture_rates}
    \end{subfigure}%
    \hfill
    \begin{subfigure}[t]{0.475\textwidth}
        \includegraphics[width=0.95\textwidth]{img/normal_mixture/type2_mean_rates.png}
        \caption{Type II error rate as a function of mean difference.}
        \label{subfig:type2_normal_mixture_mean_rate}
    \end{subfigure}%
    \caption{Measuring the Type II error rate of the considered tests on normal and normal mixture distributions as a function of sample size \cref{subfig:type2_normal_rates,subfig:type2_normal_mixture_rates} and mean differences \cref{subfig:type2_normal_mean_rate,subfig:type2_normal_mixture_mean_rate}.}\label{fig:type2-error-rates}
\end{figure}

\paragraph{Type II error rate experiments} We furthermore test the Type II error rates on samples from different distributions in \cref{fig:type2-error-rates}, sampling the score samples $500$ times for ASO and $1000$ times for the other tests from $\mathcal{N}(0.5, 1.5^2)$ and $\mathcal{N}(0, 1.5^2)$,\footnote{For the normal mixture, only the second mixture component is varied.} respectively, for a $p$-value threshold of $0.05$ and $\epsmin$ threshold of $0.2$. We see that the Type II error rate decreases with increasing sample size (\cref{subfig:type2_normal_rates,subfig:type2_normal_mixture_rates}), but is less sensitive for increasing mean difference than other tests (\cref{subfig:type2_normal_mean_rate,subfig:type2_normal_mixture_mean_rate}). Generally, we can observe the behavior to be very similar to Student's t and Mann-Whitney U test.

\paragraph{Error rates by rejection threshold} Lastly, we report the Type I and II error rates on the tested distributions using different Type I / II error rates. In \cref{table:type1-normal-tresholds,table:type1-normal-mixture-tresholds,table:type1-laplace-tresholds,table:type1-rayleigh-tresholds}, we see that ASO achieves lower error rates than other tests in almost all scenarios when faced with the fame threshold. Naturally, these thresholds cannot be interpreted the same for ASO and the other significance tests. Nevertheless, we can see that a threshold of $\tau = 0.2$ seems to roughly correspond to a $p$-value threshold of $0.05$ in terms of Type I error rate.
Type II error rates are given in \cref{table:type2-normal-tresholds,table:type2-normal-mean-tresholds,table:type2-normal-mixture-tresholds,table:type2-normal-mixture-mean-tresholds}. Here the difference between ASO and the other tests is not quite as pronounced, however, it always incurs higher error rates.

\begin{table}
    \caption{Type I error rates for samples drawn from a normal distribution as a function of sample size and different rejection thresholds.}\label{table:type1-normal-tresholds}
    \resizebox{0.975\textwidth}{!}{%
    \begin{tabular}{llrrrrrr}
    \toprule
      Sample Size & Threshold      &    ASO & Student's t & Bootstrap & Permutation & Wilcoxon & Mann-Whitney U \\
    \midrule
    5  & 0.05 &   \textbf{0.02} &       0.048 &     0.085 &       0.029 &    0.029 &          0.056 \\
       & 0.10 &  \textbf{0.034} &       0.093 &     0.149 &       0.079 &    0.088 &          0.085 \\
       & 0.20 &   \textbf{0.06} &       0.212 &     0.241 &       0.197 &     0.16 &          0.159 \\
       & 0.30 &  \textbf{0.094} &       0.299 &     0.322 &       0.286 &    0.236 &          0.284 \\
       & 0.40 &  \textbf{0.146} &       0.396 &     0.403 &        0.37 &    0.315 &          0.348 \\
       & 0.50 &  \textbf{0.216} &       0.483 &     0.483 &       0.468 &     0.49 &          0.498 \\
       \midrule
    10 & 0.05 &  \textbf{0.004} &       0.055 &     0.077 &       0.058 &    0.051 &          0.048 \\
       & 0.10 &  0.014 &       0.103 &      0.13 &        \textbf{0.11} &    0.113 &            \textbf{0.1} \\
       & 0.20 &  \textbf{0.038} &       0.196 &     0.215 &       0.201 &    0.192 &          0.194 \\
       & 0.30 &  \textbf{0.084} &       0.282 &       0.3 &       0.285 &    0.261 &          0.272 \\
       & 0.40 &  \textbf{0.138} &       0.394 &     0.398 &       0.395 &    0.387 &          0.378 \\
       & 0.50 &  \textbf{0.204} &        0.49 &     0.486 &       0.491 &    0.499 &          0.479 \\
       \midrule
    15 & 0.05 &  \textbf{0.002} &       0.059 &     0.072 &       0.057 &    0.051 &          0.052 \\
       & 0.10 &  \textbf{0.014} &       0.106 &     0.123 &       0.104 &    0.095 &          0.113 \\
       & 0.20 &  \textbf{0.042} &       0.198 &     0.215 &       0.199 &    0.186 &          0.196 \\
       & 0.30 &   \textbf{0.08} &       0.303 &     0.309 &       0.303 &    0.295 &          0.304 \\
       & 0.40 &  \textbf{0.136} &       0.395 &       0.4 &       0.392 &    0.371 &          0.368 \\
       & 0.50 &   \textbf{0.19} &       0.482 &     0.485 &       0.479 &     0.47 &          0.468 \\
       \midrule
    20 & 0.05 &  \textbf{0.004} &       0.046 &     0.058 &       0.047 &    0.043 &          0.047 \\
       & 0.10 &  \textbf{0.006} &       0.095 &     0.105 &       0.093 &    0.085 &          0.092 \\
       & 0.20 &  \textbf{0.028} &       0.181 &     0.196 &       0.177 &    0.171 &          0.183 \\
       & 0.30 &  \textbf{0.074} &        0.28 &      0.29 &       0.289 &    0.284 &          0.273 \\
       & 0.40 &   \textbf{0.12} &       0.384 &     0.389 &       0.381 &    0.372 &          0.394 \\
       & 0.50 &   \textbf{0.17} &       0.479 &     0.478 &       0.473 &    0.477 &          0.481 \\
    \bottomrule
    \end{tabular}%
    }
\end{table}

\begin{table}
    \caption{Type II error rates for samples drawn from two different normal distributions as a function of sample size and different rejection thresholds.}\label{table:type2-normal-tresholds}
    \resizebox{0.975\textwidth}{!}{%
    \begin{tabular}{lllrrrrrr}
    \toprule
      Sample Size & Threshold      &    ASO & Student's t & Bootstrap & Permutation & Wilcoxon & Mann-Whitney U \\
    \midrule
    5  & 0.05 &  0.942 &       0.883 &     \textbf{0.796} &       0.918 &    0.925 &          0.875 \\
   & 0.10 &  0.916 &       0.786 &     \textbf{0.714} &       0.802 &    0.792 &          0.819 \\
   & 0.20 &   0.87 &       0.623 &     \textbf{0.585} &       0.649 &    0.691 &          0.694 \\
   & 0.30 &  0.792 &       0.512 &      \textbf{0.48} &       0.521 &    0.597 &          0.539 \\
   & 0.40 &  0.714 &       0.399 &      \textbf{0.39} &       0.421 &    0.498 &           0.47 \\
   & 0.50 &   0.65 &       \textbf{0.302} &     0.315 &       0.318 &    0.387 &          0.391 \\
   \midrule
10 & 0.05 &  0.978 &       0.836 &     \textbf{0.791} &       0.853 &    0.864 &           0.84 \\
   & 0.10 &   0.95 &        0.73 &     \textbf{0.695} &       0.737 &    0.743 &          0.741 \\
   & 0.20 &  0.868 &        0.58 &     \textbf{0.551} &        0.58 &    0.595 &          0.576 \\
   & 0.30 &  0.802 &       0.428 &      \textbf{0.41} &       0.429 &    0.462 &          0.453 \\
   & 0.40 &  0.708 &        0.33 &     \textbf{0.328} &       0.327 &    0.347 &          0.329 \\
   & 0.50 &  0.604 &       \textbf{0.223} &     0.223 &       0.229 &    0.272 &          0.251 \\
   \midrule
15 & 0.05 &  0.984 &       0.769 &     0.734 &       0.781 &    0.788 &          0.787 \\
   & 0.10 &   0.95 &       0.643 &     \textbf{0.615} &       0.646 &    0.672 &          0.639 \\
   & 0.20 &   0.84 &        \textbf{0.47} &     0.455 &        0.48 &    0.493 &          0.481 \\
   & 0.30 &  0.716 &       0.348 &      \textbf{0.34} &        0.35 &    0.355 &          0.365 \\
   & 0.40 &   0.61 &       \textbf{0.244} &     0.245 &       0.246 &    0.276 &          0.261 \\
   & 0.50 &  0.486 &       0.177 &     0.176 &       \textbf{0.175} &    0.185 &          0.192 \\
   \midrule
20 & 0.05 &  0.976 &       0.732 &     \textbf{0.709} &       0.736 &     0.75 &          0.747 \\
   & 0.10 &  0.946 &       0.601 &     \textbf{0.586} &       0.601 &    0.614 &           0.61 \\
   & 0.20 &  0.848 &       0.406 &     \textbf{0.396} &        0.41 &    0.421 &           0.41 \\
   & 0.30 &  0.704 &       0.277 &     \textbf{0.268} &       0.272 &    0.299 &          0.289 \\
   & 0.40 &   0.58 &         \textbf{0.2} &     \textbf{0.201} &       0.198 &    0.221 &          0.206 \\
   & 0.50 &  0.444 &       \textbf{0.144} &     \textbf{0.144} &       0.147 &    0.156 &          0.152 \\
    \bottomrule
    \end{tabular}%
    }
\end{table}

\begin{table}
    \caption{Type II error rates for samples drawn from a normal distribution as a function of mean difference and different rejection thresholds.}\label{table:type2-normal-mean-tresholds}
    \resizebox{0.975\textwidth}{!}{%
    \begin{tabular}{llrrrrrr}
    \toprule
       Difference  & Treshold      &    ASO & Student's t & Bootstrap & Permutation & Wilcoxon & Mann-Whitney U \\
    \midrule
    0.25 & 0.05 &  0.984 &       0.925 &     \textbf{0.857} &       0.941 &    0.945 &           0.93 \\
     & 0.10 &  0.954 &       0.846 &     \textbf{0.781} &       0.859 &    0.844 &          0.881 \\
     & 0.20 &  0.914 &       0.705 &     \textbf{0.659} &       0.721 &    0.761 &          0.768 \\
     & 0.30 &  0.872 &       \textbf{0.585} &     0.554 &       0.606 &     0.68 &          0.622 \\
     & 0.40 &    0.8 &       0.482 &     \textbf{0.462} &       0.489 &    0.594 &          0.548 \\
     & 0.50 &  0.714 &       \textbf{0.381} &     0.387 &       0.394 &     0.48 &          0.465 \\
     \midrule
0.50 & 0.05 &  0.966 &       0.888 &     \textbf{0.805} &       0.918 &     0.92 &          0.883 \\
     & 0.10 &  0.932 &       0.784 &       \textbf{0.7} &       0.811 &    0.794 &           0.83 \\
     & 0.20 &   0.87 &       0.616 &      \textbf{0.57} &       0.652 &    0.696 &          0.698 \\
     & 0.30 &  0.812 &         0.5 &     \textbf{0.477} &       0.523 &    0.602 &          0.535 \\
     & 0.40 &  0.722 &       0.406 &     \textbf{0.397} &       0.426 &    0.505 &          0.466 \\
     & 0.50 &  0.606 &       \textbf{0.313} &     0.315 &       0.326 &    0.411 &          0.401 \\
     \midrule
0.75 & 0.05 &  0.934 &       0.822 &     \textbf{0.707} &       0.883 &    0.885 &          0.822 \\
     & 0.10 &  0.896 &       0.699 &      \textbf{0.61} &       0.725 &     0.71 &          0.764 \\
     & 0.20 &  0.798 &       0.514 &     \textbf{0.469} &       0.561 &    0.599 &          0.607 \\
     & 0.30 &  0.702 &       0.407 &      \textbf{0.37} &       0.421 &    0.515 &          0.455 \\
     & 0.40 &   0.59 &       0.308 &       \textbf{0.3} &       0.325 &    0.406 &          0.375 \\
     & 0.50 &  0.482 &       \textbf{0.223} &     \textbf{0.222} &       0.237 &    0.303 &          0.295 \\
     \midrule
1.00 & 0.05 &   0.87 &       0.739 &     \textbf{0.609} &        0.85 &     0.85 &          0.743 \\
     & 0.10 &  0.796 &       0.585 &     \textbf{0.488} &       0.678 &    0.655 &          0.659 \\
     & 0.20 &  0.712 &       0.386 &     \textbf{0.327} &       0.449 &    0.497 &          0.487 \\
     & 0.30 &   0.58 &       0.257 &     \textbf{0.232} &       0.289 &    0.388 &          0.307 \\
     & 0.40 &  0.504 &       0.178 &      \textbf{0.17} &       0.194 &    0.278 &          0.229 \\
     & 0.50 &  0.384 &       \textbf{0.115} &     \textbf{0.115} &       0.128 &    0.189 &          0.176 \\
    \bottomrule
    \end{tabular}%
    }
\end{table}

\begin{table}
    \centering
    \caption{Type I error rates for samples drawn from a normal mixture distribution as a function of sample size and different rejection thresholds.}\label{table:type1-normal-mixture-tresholds}
    \resizebox{0.975\textwidth}{!}{%
    \begin{tabular}{llrrrrrr}

    \toprule
      Sample Size & Threshold      &    ASO & Student's t & Bootstrap & Permutation & Wilcoxon & Mann-Whitney U \\
    \midrule
    5  & 0.05 &   \textbf{0.0} &         \textbf{0.0} &     0.012 &       0.028 &    0.026 &          0.003 \\
       & 0.10 &    \textbf{0.0} &       0.013 &     0.035 &       0.079 &    0.085 &          0.004 \\
       & 0.20 &    \textbf{0.0} &       0.069 &     0.104 &       0.179 &    0.153 &          0.049 \\
       & 0.30 &  \textbf{0.008} &       0.169 &     0.213 &       0.281 &    0.208 &           0.16 \\
       & 0.40 &  \textbf{0.024} &       0.338 &     0.358 &       0.363 &    0.305 &          0.244 \\
       & 0.50 &  \textbf{0.058} &       0.494 &     0.493 &       0.483 &    0.484 &          0.478 \\
       \midrule
    10 & 0.05 &    \textbf{0.0} &       0.007 &     0.018 &       0.059 &    0.049 &          0.011 \\
       & 0.10 &    \textbf{0.0} &       0.031 &      0.05 &        0.11 &    0.109 &           0.03 \\
       & 0.20 &  \textbf{0.004} &       0.102 &     0.121 &       0.205 &    0.188 &          0.109 \\
       & 0.30 &  \textbf{0.008} &       0.221 &     0.229 &       0.302 &    0.273 &          0.211 \\
       & 0.40 &  \textbf{0.034} &       0.347 &     0.349 &       0.398 &    0.379 &          0.351 \\
       & 0.50 &   \textbf{0.07} &       0.511 &     0.515 &       0.506 &    0.491 &          0.495 \\
       \midrule
    15 & 0.05 &    \textbf{0.0} &       0.006 &     0.007 &       0.055 &    0.048 &          0.004 \\
       & 0.10 &    \textbf{0.0} &       0.022 &     0.033 &       0.106 &    0.097 &          0.017 \\
       & 0.20 &  \textbf{0.002} &       0.103 &     0.118 &       0.194 &    0.202 &          0.095 \\
       & 0.30 &  \textbf{0.006} &       0.215 &      0.22 &       0.301 &    0.308 &          0.208 \\
       & 0.40 &  \textbf{0.028} &       0.356 &     0.366 &       0.415 &    0.404 &          0.328 \\
       & 0.50 &  \textbf{0.082} &       0.501 &     0.499 &       0.496 &    0.502 &          0.501 \\
       \midrule
    20 & 0.05 &    \textbf{0.0} &       0.006 &     0.007 &       0.048 &    0.045 &          0.005 \\
       & 0.10 &    \textbf{0.0} &       0.019 &     0.027 &       0.088 &    0.085 &          0.021 \\
       & 0.20 &    \textbf{0.0} &       0.104 &     0.109 &         0.2 &    0.187 &          0.097 \\
       & 0.30 &  \textbf{0.006} &       0.214 &     0.218 &       0.307 &    0.289 &          0.221 \\
       & 0.40 &  \textbf{0.032} &       0.363 &     0.369 &       0.412 &     0.39 &          0.349 \\
       & 0.50 &  \textbf{0.082} &       0.494 &     0.495 &       0.492 &    0.496 &          0.485 \\
    \bottomrule
    \end{tabular}%
    }
\end{table}

\begin{table}
    \caption{Type II error rates for samples drawn from two normal mixture distribution as a function of sample size and different rejection thresholds.}\label{table:type2-normal-mixture-tresholds}
    \resizebox{0.975\textwidth}{!}{%
    \begin{tabular}{llrrrrrr}
    \toprule
      Sample Size & Threshold      &    ASO & Student's t & Bootstrap & Permutation & Wilcoxon & Mann-Whitney U \\
    \midrule
    5  & 0.05 &    1.0 &       0.999 &     0.964 &       \textbf{0.892} &    0.897 &          0.995 \\
   & 0.10 &    1.0 &       0.962 &     0.874 &       0.728 &    \textbf{0.697} &          0.985 \\
   & 0.20 &  0.994 &       0.747 &      0.64 &       \textbf{0.474} &    0.525 &           0.87 \\
   & 0.30 &  0.976 &       0.476 &     0.422 &       \textbf{0.299} &    0.426 &          0.579 \\
   & 0.40 &  0.896 &       0.252 &     0.234 &       \textbf{0.206} &    0.326 &          0.414 \\
   & 0.50 &  0.748 &       \textbf{0.117} &     \textbf{0.118} &       0.122 &    0.222 &           0.28 \\
   \midrule
10 & 0.05 &    1.0 &       0.908 &     0.831 &       \textbf{0.552} &    0.635 &          0.926 \\
   & 0.10 &  0.996 &       0.721 &     0.641 &       \textbf{0.354} &    0.419 &           0.73 \\
   & 0.20 &  0.954 &        0.39 &     0.354 &       \textbf{0.186} &    0.247 &          0.407 \\
   & 0.30 &  0.828 &       0.191 &      0.18 &       \textbf{0.108} &    0.156 &          0.219 \\
   & 0.40 &  0.642 &       0.089 &     0.087 &       \textbf{0.068} &    0.089 &          0.107 \\
   & 0.50 &  0.452 &       0.034 &     \textbf{0.031} &       0.037 &    0.056 &          0.052 \\
   \midrule
15 & 0.05 &  0.996 &       0.829 &     0.757 &       \textbf{0.352} &    0.441 &          0.864 \\
   & 0.10 &   0.99 &       0.568 &     0.517 &       \textbf{0.213} &    0.272 &          0.628 \\
   & 0.20 &  0.928 &       0.251 &     0.234 &       \textbf{0.087} &    0.129 &          0.298 \\
   & 0.30 &  0.774 &       0.099 &     0.091 &       \textbf{0.033} &    0.058 &          0.116 \\
   & 0.40 &  0.498 &       0.027 &     0.026 &       \textbf{0.019} &    0.034 &          0.044 \\
   & 0.50 &  0.276 &       \textbf{0.009} &      \textbf{0.01} &        \textbf{0.01} &    0.013 &          0.014 \\
   \midrule
20 & 0.05 &    1.0 &       0.653 &      0.58 &       \textbf{0.204} &    0.279 &          0.666 \\
   & 0.10 &   0.98 &       0.359 &     0.333 &       \textbf{0.105} &    0.162 &          0.392 \\
   & 0.20 &  0.848 &       0.107 &     0.101 &       \textbf{0.035} &    0.064 &          0.147 \\
   & 0.30 &  0.586 &       0.038 &     0.035 &       \textbf{0.013} &    0.022 &          0.047 \\
   & 0.40 &  0.344 &        0.01 &      0.01 &       \textbf{0.008} &    0.013 &          0.017 \\
   & 0.50 &   0.13 &       \textbf{0.003} &     \textbf{0.003} &       \textbf{0.004} &    0.006 &          0.006 \\
    \bottomrule
    \end{tabular}%
    }
\end{table}

\begin{table}
    \caption{Type II error rates for samples drawn from two normal mixture distribution as a function of mean difference between two of the mixture components and different rejection thresholds.}\label{table:type2-normal-mixture-mean-tresholds}
    \resizebox{0.975\textwidth}{!}{%
    \begin{tabular}{llrrrrrr}
    \toprule
         &      &    ASO & Student's t & Bootstrap & Permutation & Wilcoxon & Mann-Whitney U \\
    diff & threshold &        &             &           &             &          &                \\
    \midrule
    0.25 & 0.05 &    1.0 &       0.997 &     0.988 &       0.958 &    0.962 &          0.998 \\
     & 0.10 &  0.998 &       0.988 &      0.96 &       0.894 &    \textbf{0.882} &          0.994 \\
     & 0.20 &  0.996 &       0.903 &     0.856 &       \textbf{0.754} &    0.792 &          0.945 \\
     & 0.30 &  0.978 &       0.762 &     0.727 &       \textbf{0.643}&    0.724 &          0.814 \\
     & 0.40 &   0.94 &       0.594 &     0.576 &        \textbf{0.53} &    0.621 &          0.704 \\
     & 0.50 &  0.886 &       \textbf{0.424} &    \textbf{0.424} &       0.444 &    0.532 &          0.563 \\
     \midrule
0.50 & 0.05 &  0.998 &       0.999 &      0.98 &       \textbf{0.931} &    0.932 &          0.996 \\
     & 0.10 &  0.998 &       0.978 &     0.931 &        0.82 &    \textbf{0.802} &           0.99 \\
     & 0.20 &  0.996 &       0.849 &     0.775 &       \textbf{0.647} &    0.695 &          0.905 \\
     & 0.30 &  0.976 &       0.659 &     0.603 &       \textbf{0.511} &    0.611 &          0.724 \\
     & 0.40 &  0.928 &       0.458 &     0.438 &       \textbf{0.407} &    0.504 &          0.577 \\
     & 0.50 &   0.84 &       \textbf{0.284} &     0.287 &        0.31 &    0.395 &          0.449 \\
     \midrule
0.75 & 0.05 &    1.0 &       0.997 &     0.966 &       \textbf{0.912} &    0.915 &          0.993 \\
     & 0.10 &  0.998 &       0.966 &     0.901 &       0.769 &    \textbf{0.746} &          0.985 \\
     & 0.20 &  0.994 &       0.802 &     0.707 &       \textbf{0.553} &    0.623 &          0.886 \\
     & 0.30 &  0.974 &       0.547 &     0.497 &       \textbf{0.397} &    0.516 &          0.651 \\
     & 0.40 &  0.922 &       0.355 &     0.337 &       \textbf{0.286} &    0.407 &          0.485 \\
     & 0.50 &  0.824 &       \textbf{0.191} &     \textbf{0.191} &       0.198 &    0.305 &          0.363 \\
     \midrule
1.00 & 0.05 &    1.0 &         1.0 &     0.961 &        \textbf{0.89} &    0.894 &          0.995 \\
     & 0.10 &    1.0 &       0.958 &     0.868 &       0.714 &    \textbf{0.682} &          0.989 \\
     & 0.20 &  0.996 &       0.715 &     0.617 &       \textbf{0.445} &    0.505 &          0.872 \\
     & 0.30 &  0.962 &       0.432 &      0.38 &       \textbf{0.291} &    0.419 &          0.545 \\
     & 0.40 &   0.87 &       0.253 &     0.235 &       \textbf{0.204} &    0.308 &          0.408 \\
     & 0.50 &  0.702 &        \textbf{0.12} &      \textbf{0.12} &       0.132 &    0.208 &          0.263 \\
    \bottomrule
    \end{tabular}%
    }
\end{table}

\begin{table}
    \caption{Type I error rates for samples drawn from a Laplace distribution as a function of sample size and different rejection thresholds.}\label{table:type1-laplace-tresholds}
    \resizebox{0.975\textwidth}{!}{%
    \begin{tabular}{llrrrrrr}
    \toprule
      Sample Size & Threshold      &    ASO & Student's t & Bootstrap & Permutation & Wilcoxon & Mann-Whitney U \\
    \midrule
    5  & 0.05 &  \textbf{0.022} &       0.053 &      0.11 &       0.048 &    0.046 &          0.066 \\
       & 0.10 &  \textbf{0.038} &       0.117 &     0.164 &       0.106 &    0.116 &          0.097 \\
       & 0.20 &  \textbf{0.088} &       0.223 &     0.261 &       0.208 &    0.187 &          0.169 \\
       & 0.30 &  \textbf{0.124} &       0.319 &     0.343 &       0.295 &    0.234 &          0.286 \\
       & 0.40 &  \textbf{0.154} &       0.427 &     0.445 &       0.398 &    0.322 &          0.379 \\
       & 0.50 &  \textbf{0.218} &       0.509 &      0.51 &       0.491 &    0.506 &          0.508 \\
       \midrule
    10 & 0.05 &  \textbf{0.004} &       0.059 &     0.077 &        0.06 &    0.046 &          0.051 \\
       & 0.10 &  \textbf{0.012} &       0.114 &     0.142 &       0.111 &    0.106 &          0.098 \\
       & 0.20 &  \textbf{0.056} &       0.218 &     0.236 &       0.216 &    0.202 &          0.199 \\
       & 0.30 &  \textbf{0.104}&       0.314 &      0.33 &       0.318 &     0.29 &          0.291 \\
       & 0.40 &  \textbf{0.164} &       0.404 &     0.407 &       0.398 &    0.378 &            0.4 \\
       & 0.50 &  \textbf{0.238} &       0.475 &     0.475 &       0.473 &    0.481 &          0.486 \\
       \midrule
    15 & 0.05 &    \textbf{0.0} &       0.052 &     0.066 &       0.048 &    0.048 &          0.048 \\
       & 0.10 &  \textbf{0.012} &         0.1 &     0.117 &       0.103 &      0.1 &          0.101 \\
       & 0.20 &  \textbf{0.028} &       0.204 &      0.22 &       0.199 &    0.199 &          0.187 \\
       & 0.30 &   \textbf{0.07} &       0.311 &     0.319 &       0.303 &    0.296 &          0.294 \\
       & 0.40 &   \textbf{0.12} &       0.404 &     0.409 &       0.402 &    0.378 &          0.394 \\
       & 0.50 &  \textbf{0.194} &        0.51 &     0.514 &       0.511 &    0.504 &          0.519 \\
       \midrule
    20 & 0.05 &  \textbf{0.004} &       0.044 &     0.047 &       0.048 &    0.057 &          0.052 \\
       & 0.10 &   \textbf{0.01} &       \textbf{0.099} &     0.113 &       0.104 &    0.103 &          0.101 \\
       & 0.20 &   \textbf{0.03} &       0.214 &     0.232 &       0.215 &    0.199 &          0.202 \\
       & 0.30 &  \textbf{0.064} &       0.312 &     0.325 &       0.308 &    0.297 &          0.307 \\
       & 0.40 &  \textbf{0.138} &       0.414 &     0.413 &       0.415 &    0.381 &          0.405 \\
       & 0.50 &   \textbf{0.22} &       0.507 &     0.505 &       0.501 &    0.485 &          0.496 \\
    \bottomrule
    \end{tabular}%
    }
\end{table}

\begin{table}
    \caption{Type I error rates for samples drawn from a Rayleigh distribution as a function of sample size and different rejection thresholds.}\label{table:type1-rayleigh-tresholds}
    \resizebox{0.975\textwidth}{!}{%
    \begin{tabular}{llrrrrrr}
    \toprule
      Sample Size & Threshold      &    ASO & Student's t & Bootstrap & Permutation & Wilcoxon & Mann-Whitney U \\
    \midrule
    5  & 0.05 &  \textbf{0.012} &       0.054 &     0.107 &       0.028 &    0.028 &          0.054 \\
       & 0.10 &  \textbf{0.034} &       0.108 &     0.147 &       0.089 &    0.096 &          0.088 \\
       & 0.20 &  \textbf{0.076} &       0.203 &     0.235 &       0.187 &    0.162 &          0.165 \\
       & 0.30 &   \textbf{0.11} &       0.319 &     0.342 &       0.302 &    0.229 &          0.291 \\
       & 0.40 &  \textbf{0.146} &       0.423 &     0.435 &       0.415 &    0.331 &           0.36 \\
       & 0.50 &  \textbf{0.198} &       0.532 &     0.539 &       0.523 &     0.53 &          0.524 \\
       \midrule
    10 & 0.05 &  \textbf{0.012} &       0.046 &     0.062 &       0.043 &    0.039 &          0.041 \\
       & 0.10 &  \textbf{0.018} &       0.087 &     0.107 &       0.093 &    0.094 &          0.084 \\
       & 0.20 &  \textbf{0.044} &       0.187 &     0.206 &        0.18 &    0.172 &          0.187 \\
       & 0.30 &  \textbf{0.064} &       0.295 &     0.314 &       0.297 &    0.265 &          0.284 \\
       & 0.40 &  \textbf{0.114} &       0.401 &     0.405 &       0.399 &    0.373 &          0.412 \\
       & 0.50 &   \textbf{0.18} &       0.507 &     0.514 &       0.505 &      0.5 &          0.508 \\
       \midrule
    15 & 0.05 &  \textbf{0.004} &        0.05 &     0.064 &       0.049 &     0.05 &          0.054 \\
       & 0.10 &   \textbf{0.01} &         0.1 &     0.115 &         0.1 &    0.103 &          0.104 \\
       & 0.20 &  \textbf{0.036} &       0.194 &     0.201 &       0.182 &    0.187 &          0.187 \\
       & 0.30 &   \textbf{0.07} &       0.295 &     0.302 &       0.287 &    0.294 &          0.291 \\
       & 0.40 &  \textbf{0.114} &       0.386 &     0.394 &       0.379 &    0.371 &          0.373 \\
       & 0.50 &  \textbf{0.198} &       0.481 &     0.484 &       0.487 &    0.472 &          0.497 \\
       \midrule
    20 & 0.05 &  \textbf{0.002} &       0.054 &     0.064 &       0.059 &    0.055 &          0.052 \\
       & 0.10 &  \textbf{0.004} &       0.115 &     0.121 &       0.113 &    0.103 &          0.113 \\
       & 0.20 &   \textbf{0.03} &       0.195 &     0.205 &       0.202 &    0.187 &          0.204 \\
       & 0.30 &  \textbf{0.058} &       0.281 &     0.287 &       0.277 &    0.283 &          0.291 \\
       & 0.40 &   \textbf{0.13} &       0.377 &     0.386 &       0.375 &    0.368 &          0.384 \\
       & 0.50 &   \textbf{0.19} &       0.489 &     0.493 &       0.493 &    0.468 &          0.469 \\
    \bottomrule
    \end{tabular}%
    }
\end{table}

\end{document}